\documentclass[letterpaper]{article} % DO NOT CHANGE THIS
\usepackage{aaai25}  % DO NOT CHANGE THIS
\usepackage[dvipsnames]{xcolor}
\usepackage{xcolor}
\usepackage{times}  % DO NOT CHANGE THIS
\usepackage{helvet}  % DO NOT CHANGE THIS
\usepackage{courier}  % DO NOT CHANGE THIS
\usepackage[hyphens]{url}  % DO NOT CHANGE THIS
\usepackage{graphicx} % DO NOT CHANGE THIS
\usepackage{natbib}  % DO NOT CHANGE THIS AND DO NOT ADD ANY OPTIONS TO IT
\usepackage{caption} % DO NOT CHANGE THIS AND DO NOT ADD ANY OPTIONS TO IT
\usepackage{algorithm}
\usepackage{algorithmic}
\usepackage{pdfpages}
\usepackage{newfloat}
\usepackage{listings}
\DeclareCaptionStyle{ruled}{labelfont=normalfont,labelsep=colon,strut=off} % DO NOT CHANGE THIS
\floatstyle{ruled}
\newfloat{listing}{tb}{lst}{}
\floatname{listing}{Listing}
\title{VerilogCoder: Autonomous Verilog Coding Agents with Graph-based Planning and Abstract Syntax Tree (AST)-based Waveform Tracing Tool}
\author {
    Chia-Tung Ho,
    Haoxing Ren,
    Brucek Khailany
}
\affiliations {
    NVIDIA Research\\
    chiatungh@nvidia.com, 
    haoxingr@nvidia.com, bkhailany@nvidia.com
}

\usepackage{bibentry}
\begin{document}

\maketitle

\begin{abstract}
Due to the growing complexity of modern Integrated Circuits (ICs), automating hardware design can prevent a significant amount of human error from the engineering process and result in less errors. Verilog is a popular hardware description language for designing and modeling digital systems; thus, Verilog generation is one of the emerging areas of research to facilitate the design process. In this work, we propose VerilogCoder, a system of multiple Artificial Intelligence (AI) agents for Verilog code generation, to autonomously write Verilog code and fix syntax and functional errors using collaborative Verilog tools (i.e., syntax checker, simulator, and waveform tracer). Firstly, we propose a task planner that utilizes a novel Task and Circuit Relation Graph retrieval method to construct a holistic plan based on module descriptions. To debug and fix functional errors, we develop a novel and efficient abstract syntax tree (AST)-based waveform tracing tool, which is integrated within the autonomous Verilog completion flow. The proposed methodology successfully generates 94.2\% syntactically and functionally correct Verilog code, surpassing the state-of-the-art methods by 33.9\% on the VerilogEval-Human v2 benchmark. 
%\textcolor{black}{The code and data can be found at \em{https://github.com/NVlabs/VerilogCoder}.}
\end{abstract}

\begin{links}
     \link{Code}{https://github.com/NVlabs/VerilogCoder}
\end{links}

\section{Introduction}
Designing modern integrated circuits requires designers to write code in hardware description languages such as Verilog and VHDL to specify hardware architectures and model the behaviors of digital systems.
Due to the growing complexity of VLSI design, writing Verilog and \textcolor{black}{VHDL} is time-consuming and prone to bugs, necessitating multiple iterations for debugging functional correctness.
Consequently, reducing design costs and designer effort for completing hardware specifications has emerged as a critical need.

\textcolor{black}{Large Language Models (LLMs) have} shown remarkable capacity to comprehend and generate natural language at a massive scale, leading to many potential applications and benefits across various domains.
In the field of coding, LLM can assist developers by suggesting code snippets, offering solutions to fix bugs, and even generating the code with explanation~\cite{mastropaolo2023robustness, nijkamp2023codegen2}.
Several works have focused on refining LLMs with selected datasets for Verilog generation~\cite{liu2023verilogeval, thakur2024verigen}.
Pei {\em et al.}~\cite{pei2024betterv} proposed leveraging instruct-tuned LLM and a generative discriminators to optimize Verilog implementation with the considerations of PPA (Power, Performance, Area). However, these works lack of a mechanism to fix syntactic or functional errors, thus, they still struggle to generate functionally correct Verilog code.
Recently, Tsai {\em et al.}~\cite{tsai2023rtlfixer} presented an autonomous agent framework incorporating feedback from simulators and Retrieval Augmented Generation to fix syntax errors, but it failed to improve the functional success rate. 

In this work, we propose a framework leveraging multiple Artificial Intelligence (AI) agents for Verilog code generation, which autonomously writes the Verilog code and fixes syntax and functional errors using collaborative Verilog toolkits and \textcolor{black}{the} ReAct~\cite{yao2022react} technique. 
In the framework, we develop a novel task planner to generate high-quality plans, and integrate a crafted Abstract Syntax Tree (AST)-based waveform tracing tool for improving the functional success rate. Our contributions are as follows.

\begin{itemize}
    \item We are the first to explore the use of mult-AI agents for autonomous Verilog code completion, including syntax correction, and functional correction. 
    \item We have developed a novel \textcolor{black}{Task and Circuit Relation Graph (TCRG)} based task planner to create a high-quality plan with step-by-step sub-tasks and related circuit information (i.e., signal, signal transition, and single examples). % These sub-tasks are then executed by designated agents autonomously and sequentially.
    \item We propose a novel Abstract Syntax Tree (AST)-based waveform tracing tool to assist the LLM agent in fixing functional correctness.
    \item We conduct extensive and holistic ablation studies of each key component 
    % (i.e., task planner, waveform debugging tools) 
    on the VerilogEval-Human v2 benchmark~\cite{Batten2024verilogeval2}. We demonstrate the proposed VerilogCoder achieve 94.2\% pass rate, including syntax and functional correctness, and outperform the one of the state-of-the-art methods by 33.9\%.
\end{itemize}

\begin{figure*} [!t]
	\centering
	\includegraphics[width=1.88\columnwidth]{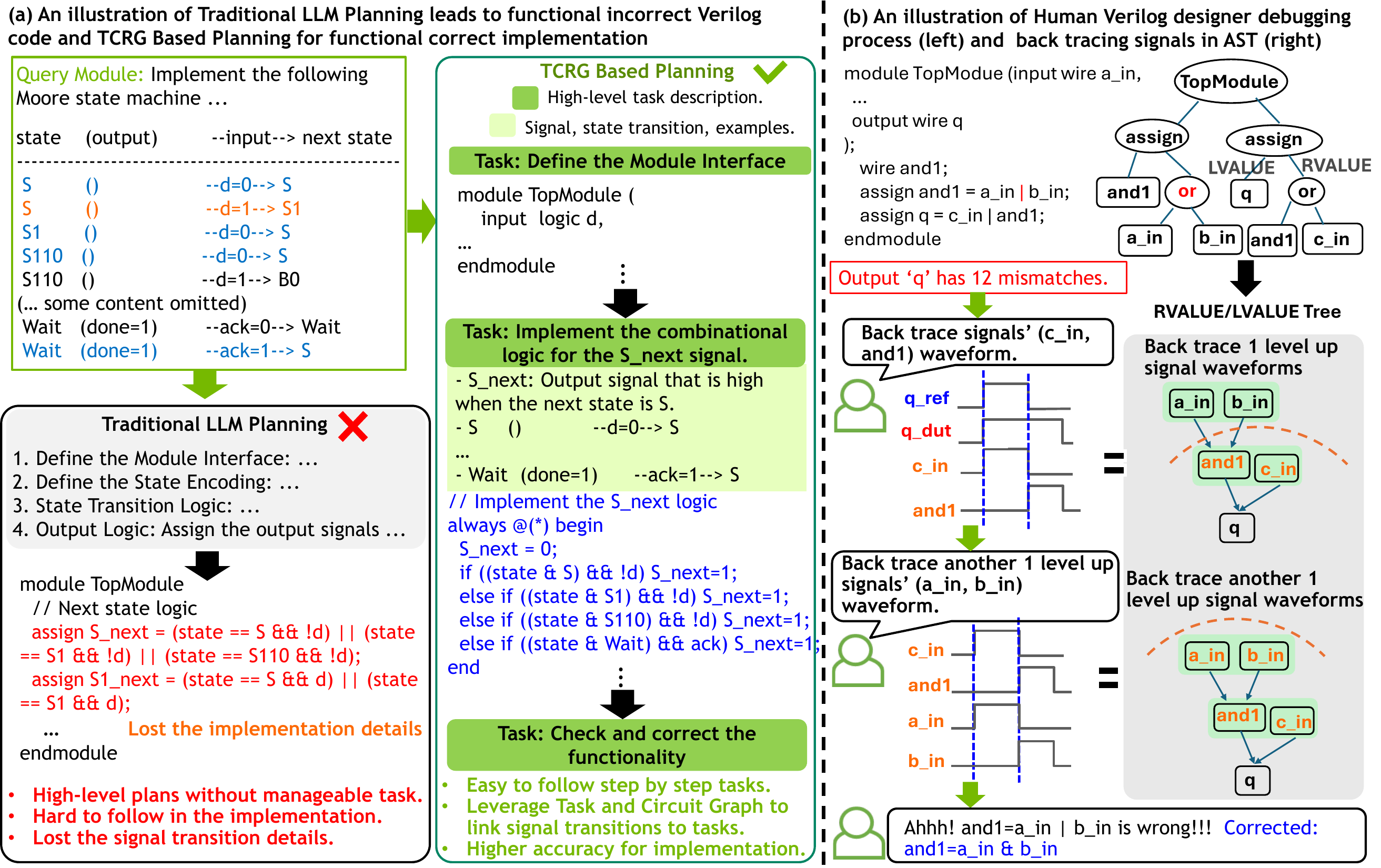}
	% \vspace{-0.3cm}
    \caption{Illustrations of (a) traditional LLM planning versus TCRG based planning, and (b) human Verilog designer debugging process and AST signal back tracing in Motivation and Preliminary Study section.}
	\label{IllustrationChanllengesFig}
    % \vspace{-0.6cm}
\end{figure*}

The remaining sections are organized as follows. \textcolor{black}{We first review prior works on AI agents and multi-AI agent systems.}
Then, we introduce and describe our novel VerilogCoder in details. Lastly, we present main \textcolor{black}{experimental results} and conclude the paper.

\section{Background} \label{BackgroundSection}
Autonomous agents have long been a research focus in academic and industrial communities across various fields. 
Recently, LLMs have shown great potential of human-level intelligence through the acquisition of vast amounts of knowledge, documents and textbooks, leading to a surge in research on LLM-based autonomous agents. 
Here, we firstly review prior AI agent works and introduce the multi-AI agent frameworks below.

\begin{figure*} [!t]
	\centering
	\includegraphics[width=1.88\columnwidth]{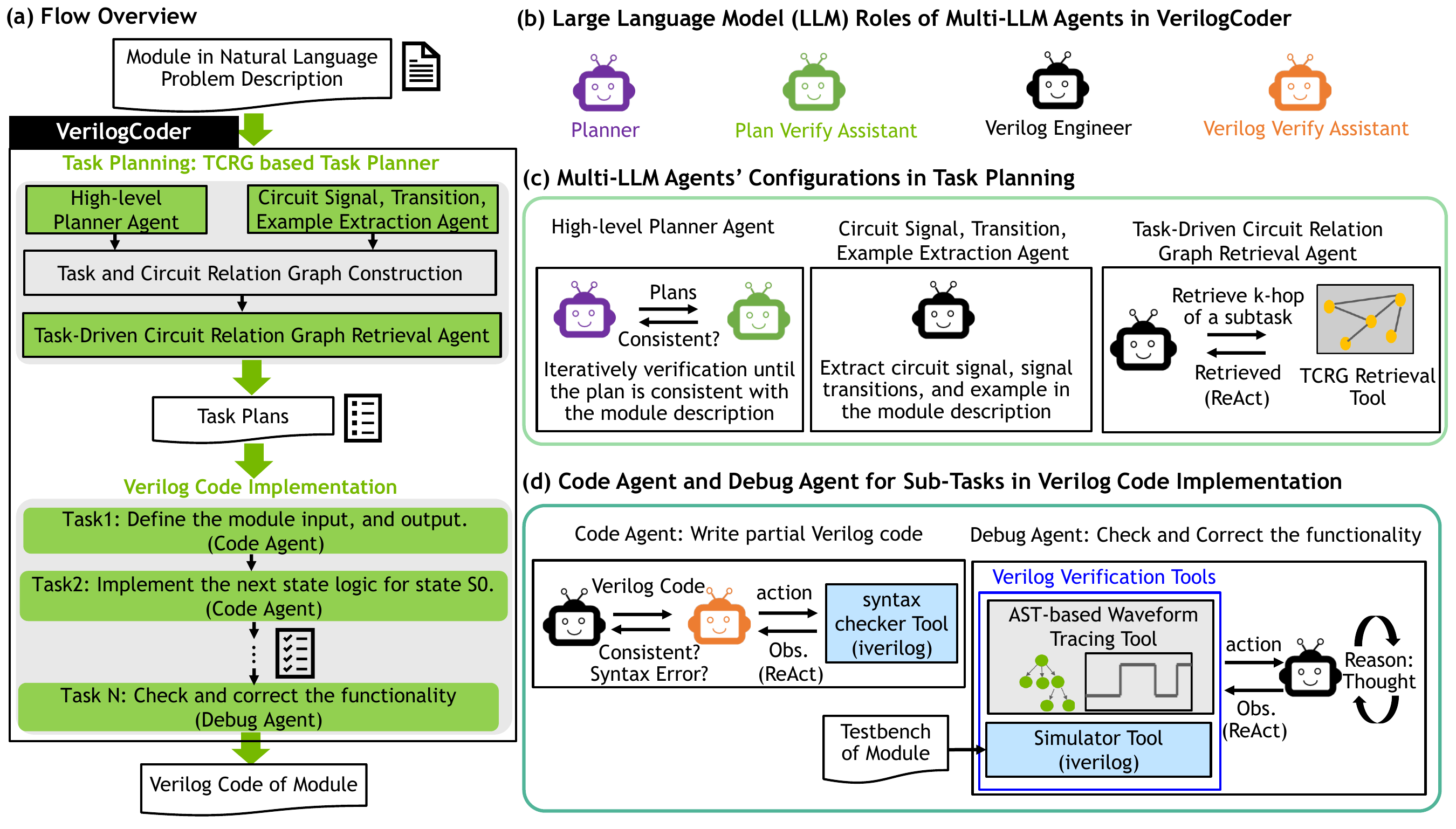}
    % \vspace{-0.3cm}
	\caption{Flow overview of VerilogCoder. (a) Overall flow for Verilog code implementation task. (b) LLM roles of multi-LLM agents. (c) Multi-LLM agents in Task Planning. (d) Multi-LLM Agents for sub-tasks in Verilog Code Implementation.}
	\label{FlowOverviewFig}
    % \vspace{-0.6cm}
\end{figure*}

\subsection{AI Agent}
Several works study the architecture of LLM-based autonomous \textcolor{black}{agents} to effectively perform diverse tasks~\cite{wang2024survey, weng2023agent}. 
From these studies, an LLM-powered autonomous agent system is composed of several key components: 
(a) Planning, (b) Memory, (c) Action, etc. 
The planning module enables the agent to break down large tasks into smaller, manageable sub plans, enabling efficient handling of complex tasks. 
In the memory module, short-term memory consists of chat history and in-context learning techniques to guide LLM actions. 
Long-term memory consolidates important information over time and provides the agent with the capability to retain and recall it over extended periods.
The action module translates the agent's decisions into outcomes for solving tasks.
The actions of an autonomous LLM-based agent can be categorized into two classes: (1) External tools for additional information and the expansion of the agent's capabilities, and (2) Internal knowledge of the LLMs, such as summary, conversation, etc.

Recently, AI agents empowered by LLMs (i.e., OpenDevin~\cite{opendevin2024}, SWE-agent~\cite{yang2024swe}, AgentCoder~\cite{huang2023agentcoder}, etc) have shown impressive performance in software engineering for solving real world challenging benchmarks (i.e., SWE-Bench, HumanEval) 
through planning, memory management, actions involving external environment tools.

\subsection{Multi-AI Agents}
In addition to single AI agents, many researchers are starting to explore the capabilities of multiple AI agents for solving complex tasks.
Autogen~\cite{wu2023autogen} \textcolor{black}{has} been proposed to enable multiple agents to operate in various modes (i.e., hierarchical chat, multi-agent conversation, etc.) that employ combinations of LLMs, human inputs, and tools.
crewAI~\cite{crewAI2024} facilitates process-oriented solving with a crew of customized multi-AI agents operating as a cohesive unit.
Currently, the applications of these multi-AI agent frameworks are mostly for general tasks (i.e., QA, summarization, coding copilot, etc.).

However, these agent frameworks cannot be directly used for designing hardware because solving hardware tasks requires integrated domain knowledge and specific hardware design toolkits (i.e., circuit simulators, waveform debugging tools) to analyze signals, trace signal transitions, and decompose tasks into manageable sub-tasks from circuit architecture and signal transaction perspectives.
% Here, we first built our VerilogCoder on top of Autogen to leverage its flexible and customized multi-AI agent capabilities.
% Then, we developed a novel TCRG-based task planner to divide tasks into manageable sub-tasks with a high-level hardware architecture view, lower-level state changes, and signal transition perspectives.
% Moreover, we designed an AST-based waveform tracing tool to improve the hardware debugging abilities (i.e., functional correctness) of our autonomous VerilogCoder.

\section{Motivation and Preliminary Study} \label{MotivationSection}
\textcolor{black}{Given a hardware module description, hardware designers usually \textcolor{black}{write Verilog using} the following steps: (1) decompose the task into manageable sub-tasks, (2) implement Verilog code for each sub-task, and (3) \textcolor{black}{iterate between Verilog simulations, signal waveform debugging, and code updates until all output signals match expected behavior.}
It is very challenging to autonomously complete a functionally correct Verilog module using LLM agents since it requires domain knowledge to break down the task into meaningful sub-tasks and comprehend the hardware descriptions and waveform during the functional debug process.}
\textcolor{black}{Consequently, we first discuss the issues of using traditional LLM planning on writing Verilog code of a Finite State Machine (FSM) module. 
Then, we study the functional debug process of \textcolor{black}{a Verilog module and propose a debugging tool that enables} LLM agents to autonomously correct the functional errors.}

\subsection{Planning} 
Planning is one of the core modules for an agent~\cite{wang2024survey, weng2023agent} to decompose a complex task into manageable sub-tasks.
%Traditional planning would leverage \textcolor{black}{a} LLM to analyze the task and decompose the complex task into manageable sub-tasks. 
For Verilog coding, the traditional LLM-generated plans usually lack of the details of relevant signals, and signal transitions for each sub-task, thus, leading to incorrect functionality implementation of Verilog \textcolor{black}{modules}. 
Figure~\ref{IllustrationChanllengesFig}(a) shows an illustration of using the traditional LLM and TCRG based planning methods on a FSM module. 
The implementation of traditional LLM planning lost part of \textcolor{black}{the state transitions} for S\_next, and S1\_next signals, thus, leading to an incorrect FSM module. Therefore, it is important to guide the agent to implement each sub-task step by step with essential signals, and state transition information. Once the state transition information and signal definitions are included with the sub-task plan, LLM can generate the correct code. Signals and state transition information can be extracted from the problem descriptions. 
In this work, we structure sub-task, signal, and state transition information in a graph format and call it the \textcolor{black}{TCRG}. 
Consequently, we study the benefits of leveraging the \textcolor{black}{TCRG} to assist the planning to generate sub-tasks that include not only high-level task goals but also the signal, and signal transition information to complete functional correct Verilog module.

\begin{figure*} [!t]
	\centering
	\includegraphics[width=1.7\columnwidth]{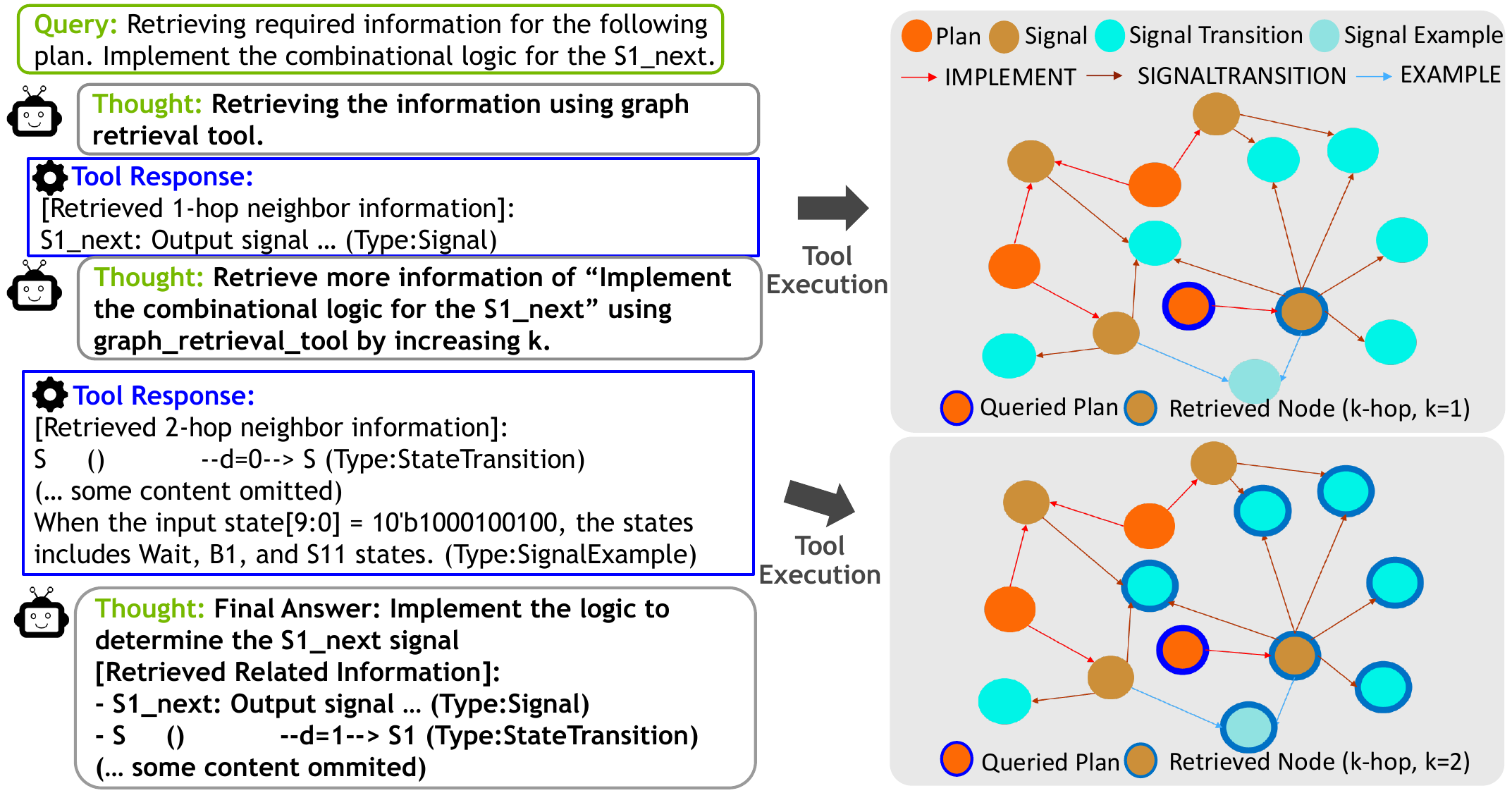}
	% \vspace{-0.3cm}
    \caption{An illustration of task-driven circuit relation graph retrieval agent reasoning and interacting with the developed {\em TCRG retrieval tool} to enrich the task with the relevant circuit and signal descriptions.}
	\label{PlanDrivenGraphExampleFig}
    % \vspace{-0.7cm}
\end{figure*}

\subsection{Functional Debug with Waveform} 
Figure~\ref{IllustrationChanllengesFig}(b) shows \textcolor{black}{a typical functional debug process for a human Verilog designer.}
Given the mismatched signals, \textcolor{black}{a human Verilog designer} traces the signals and their waveform iteratively until they know how to fix the functionality. 
This backtracing procedure is the same as tracing the RVALUE of the target signals in the AST. 
\textcolor{black}{Inspired by the human Verilog designer debug process}, we propose to incorporate the hardware signal structure, and waveform, to assist LLM agents in fixing functional errors of the generated Verilog module. This process can be implemented with a tool based on AST and waveform tracing. 
\textcolor{black}{Several prior works~\cite{alon2019code2vec, bairi2024codeplan, bui2023codetf} developed AST-based methods/tools (i.e., encoded AST paths, AST dependency graph, etc) to assist LLM in capturing structural information from the code for improving the capabilities of LLMs on various software engineering tasks such as code classification, understanding, and code completion. Here, the use of AST for signal tracing in our work is novel.}

\section{VerilogCoder} \label{VerilogCodeAgentsSection}
We introduce the details of VerilogCoder, which consist of a task planning and Verilog code implementation. The multi-AI agents of VerilogCoder operate with developed \textcolor{black}{TCRG retrieval and Verilog tools} through \textcolor{black}{the ReAct~\cite{yao2022react}} technique in a cohesive and orchestrated manner. 
% We enable the LLM-empowered multi-AI agents to autonomously decompose the Verilog coding task into a list of manageable sub-tasks considering the circuit architecture and functionality perspectives, and follow the plan to solve each sub-task. For agent memory, we keep original query and the last six chats in the chat history.

\subsection{Flow Overview}
We outline the overall flow of VerilogCoder in Figure~\ref{FlowOverviewFig}(a). Given the natural language problem description of a module~\cite{Batten2024verilogeval2}, the novel Task and Circuit Relation Graph (TCRG) based task planner first generates the task plans. 
Then, a task dependency graph is built according to the task plans and \textcolor{black}{its sub-tasks are assigned to Multi-LLM agents that write Verilog code and correct the functionality using a collaborative Verilog toolkit (i.e., syntax checker, simulator, and the proposed novel AST-based waveform tracing tool).} 
\textcolor{black}{In the flow, each agent may consist of multiple LLMs with different roles, which are listed in Figure~\ref{FlowOverviewFig}(b), to complete each step correctly and consistently.
Some of the agents are equipped with provided TCRG and Verilog tools to reason and act through Thought, Action, and Observation tracing of the ReAct prompting mechanism~\cite{yao2022react}.
For agent memory, we keep the original query and the last four chats in the chat history.}
The corresponding testbench of the module is used only for running Verilog simulator to check the functional correctness.

\begin{figure*} [!t]
	\centering
	\includegraphics[width=1.88\columnwidth]{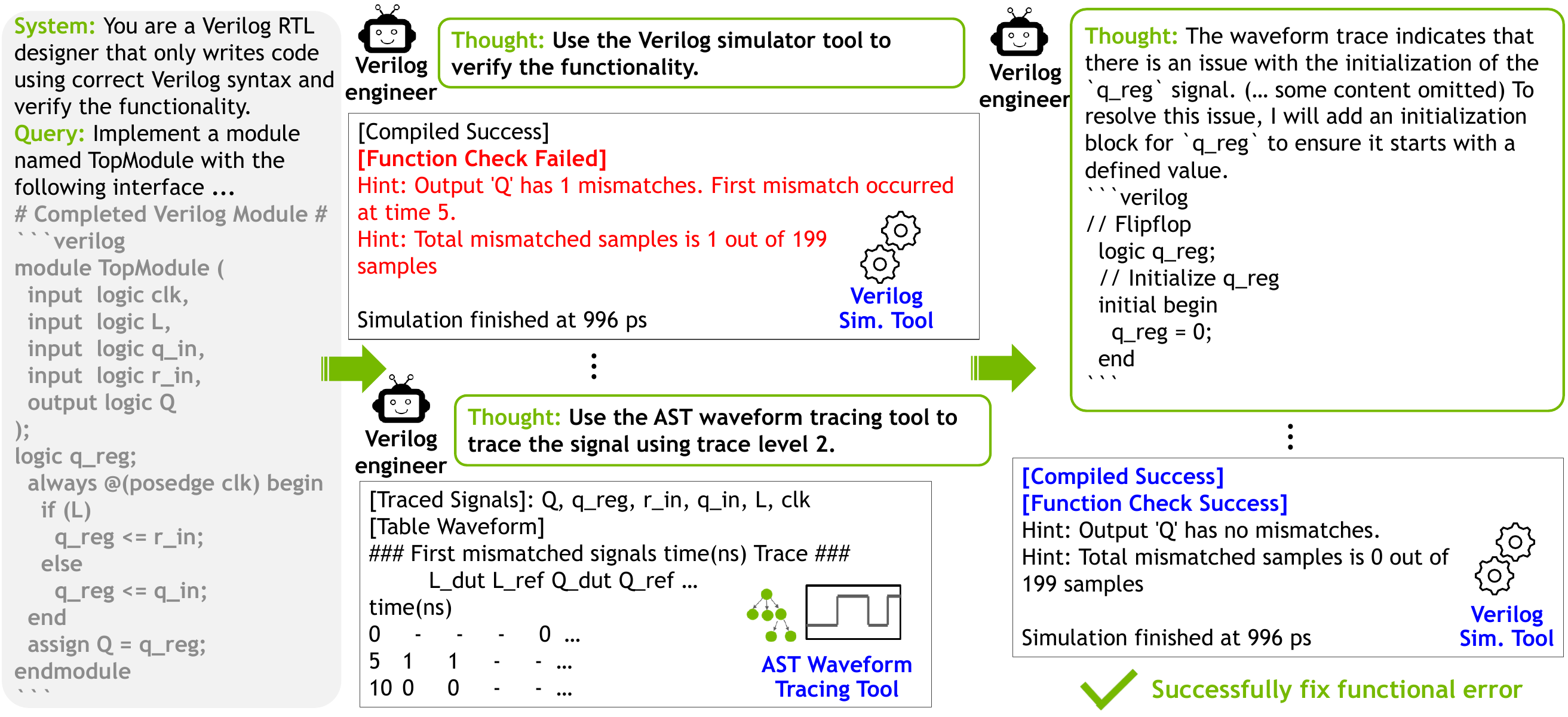}
        % \vspace{-0.3cm}
	\caption{An example of Debug Agent reasoning and interacting with simulator and AST-based waveform tracing tool.}
	\label{TaskOrientedMultiAIExampleFig}
    % \vspace{-0.6cm}
\end{figure*}

\subsection{Task Planning}
We introduce a novel and effective \textcolor{black}{TCRG} based Task Planner that constructs a high-quality plan encompassing not only the high-level objectives but also the relevant descriptions or definitions of signals, signal transitions, and examples for each sub-task. 
\textcolor{black}{Recently, many works have utilized large language models (LLMs) to analyze texts and extract entities and relations for knowledge graph construction~\cite{edge2024local, kommineni2024human, zhang2024extract}.
Inspired by these works, we leverage LLM agents to construct the \textcolor{black}{TCRG} with designer guidelines.}
In Figure~\ref{FlowOverviewFig}(a), the task plan generation flow comprises four components: (1) High-level planner agent, (2) Circuit signal, transition, example extraction agent, (3) \textcolor{black}{TCRG} construction, and (4) Task-driven circuit relation graph retrieval agent.
Figure~\ref{FlowOverviewFig}(c) shows the configuration and tools of each AI agent in TCRG based Task Planner. 

\subsubsection{High-level planner agent}
The high-level planner agent consists of a planner and a plan verification assistant, as shown in Figure~\ref{FlowOverviewFig}(c).
Given the module description or specification, the planner first decomposes the task into sub-tasks, which mostly consist of high-level task descriptions. Then, the plan verification assistant checks the consistency between the sub-tasks and the module description, providing suggestions to modify the plan if any inconsistencies are found. This iterative process continues until the planner's plan is verified to be consistent with the module description.

\subsubsection{Circuit signal, transition, example extraction agent}
A LLM acts as a Verilog engineer, extracting circuit signals, transitions, and examples from the given module description or specification into JSON format, as shown in Figure~\ref{FlowOverviewFig}(c).
The extracted information is represented as nodes in the subsequent \textcolor{black}{TCRG} construction. 
\textcolor{black}{The examples of extracted signals, transitions, and signal examples are "w: input signal examined by FSM in state B", "State A to State B: FSM moves to state B when s = 1.", and "For example, when the input w = 1, 1, 0 in these three clock cycles, output z is 
set to 1 for the following cycle.", respectively.}

\subsubsection{TCRG construction}
We create nodes from the previously generated high-level task descriptions, extracted circuit signals, transitions, and examples. 
\textcolor{black}{We then sequentially create the relations (edges) between nodes: task nodes to signal nodes, signal nodes to transition nodes, and signal nodes to example nodes, using "IMPLEMENTS", "SIGNALTRANSITION", and "EXAMPLES" relationships, respectively.}
% We then use an LLM to create edges between each pair of nodes, defining relationships as "IMPLEMENTS" "SIGNALTRANSITION" and "EXAMPLES".

\subsubsection{Task-driven circuit relation graph retrieval agent}
Here, an LLM (acting as a Verilog Engineer) autonomously retrieves relevant signal and circuit descriptions and compiles this information for each sub-task using the collaborative {\em TCRG retrieval tool} through Thought-Action-Observation ReAct tracing~\cite{yao2022react}, as shown in Figure~\ref{FlowOverviewFig}(c). 
We firstly introduce the tool and then describe the workflow of the retrieval agent.

\textcolor{black}{{\em TCRG retrieval tool} assists the task-driven circuit relation graph retrieval agent in obtaining relevant descriptions or definitions of signals, signal transitions, and examples related to a specified sub-task in the constructed \textcolor{black}{TCRG}. The inputs are the sub-task description in string format and an integer value, k, which indicates the number of hops for retrieval from the sub-task node in the graph. Here, k is determined by the AI agent automatically through the Thought-Action-Observation reasoning trace. The output consists of the retrieved k-hop signals, signal transitions, and examples corresponding to the sub-task node.}

The retrieval agent reasons and interacts with the {\em TCRG retrieval tool} to incorporate additional information  as illustrated in Figure~\ref{PlanDrivenGraphExampleFig}. Ultimately, the retrieval agent compiles the retrieved circuit and signal information from the graph and removes irrelevant information from the final answer.

\subsection{Verilog Code Implementation}
\textcolor{black}{We describe the Verilog code implementation flow of writing Verilog code and ensuring the functionality of the written Verilog module in detail.} 
Given a task plan, the task dependency graph is created. 
A child task can not be executed until all its parent tasks have been completed without errors.
The sub-tasks are divided into two types: (1) {\em Type1}: Writing Verilog code for partial function/logic, and (2) {\em Type2}: Verifying and debugging the generated Verilog module. 
The code agent and debug agent are assigned to complete the {\em Type1} sub-task and {\em Type 2} sub-task, respectively.
We first discuss the Verilog tools including a third-party simulator (i.e., iverilog~\cite{williams2002icarus}) and customized {\em AST-based waveform tracing tool}. Then, we introduce \textcolor{black}{a code agent and a debug agent}.

%\begin{figure} [!t]
%	\centering
%	\includegraphics[width=1\columnwidth]{./Figures/ASTWaveTraceOutputExample_v4.pdf}
%    \vspace{-0.7cm}
%	\caption{An example of the response of AST-based waveform tracing tool with back tracing level=2.}
%	\label{ASTTracerOutputExampleFig}
%    \vspace{-0.5cm}
%\end{figure}

\subsubsection{\textcolor{black}{Verilog Tools}}
The Verilog tools to assist agents for code implementation are listed below.

\noindent {\em Syntax checker tool}: We use iverilog to compile the generated Verilog code module and provide compiled messages as feedback for syntax checking. 

\noindent {\em Verilog simulator tool}: We use iverilog to compile the generated Verilog code module and launch the Verilog simulation. If the generated Verilog code module contains syntax errors, the tool reports the lines where these errors occur. On the other hand, the tool also reports the simulation results, including the number of mismatches in output signals and the first mismatched time point. Additionally, the tool generates a VCD file format for waveform tracing.

\noindent {\em AST-based waveform tracing tool (AST-WT)}: We developed a novel AST-based waveform tracing tool to assist agents in back-tracing the waveform of signals from mismatched output signals.
Here, we extract the AST of generated Verilog module using Pyverilog library~\cite{Takamaeda:2015:ARC:Pyverilog}. 
By inputting the mismatched output signals from the Verilog simulation tool and the desired back-tracing level, the tool starts from the mismatched signal and iteratively extracts the RVALUE signals until it reaches the specified back-tracing level in the AST, as the illustration shown in Figure~\ref{IllustrationChanllengesFig}(b).
The back-tracing level parameter is determined 
dynamically by the AI agent through the Thought-Action-Observation reasoning trace. The output includes the Verilog code reference, a tabular waveform of the mismatched signal, and the extracted RVALUE signals.
% Figure~\ref{ASTTracerOutputExampleFig} shows an example of the developed AST-based waveform tracing tool's response. 
% In this example, the debug agent call the AST-based waveform tracing tool with the simulator output (i.e., a mismatched done signal), and a back-tracing level (i.e., 2). The tool responses the reference Verilog code and the tabular signal waveform including extracted RVALUE signals (i.e., state, state$\_$next, and counter).

\subsubsection{Code Agent} 
For the code agent to write syntax-correct and consistent Verilog code, there are two LLMs: one acting as a Verilog Engineer and the other as a Verilog Verification Assistant, as shown in Figure~\ref{FlowOverviewFig}(d). 
The Verilog Engineer writes the Verilog code according to the sub-task, while the Verilog Verification Assistant ensures that the written Verilog code is consistent with the sub-task requirements and free of syntax errors using the {\em syntax checker tool}.
If there are syntax errors or inconsistencies between the written Verilog code and the sub-task description, the Verilog Verification Assistant will provide suggestions to the Verilog Engineer for fixing the issues. 
This process continues iteratively between the Verilog Engineer and the Verilog Verification Assistant until the generated Verilog code is free of syntax errors and consistent with the sub-task description.  
% Figure~\ref{TaskOrientedMultiAIExampleFig}(a) shows an example of the iterative process between Verilog engineer and Verilog verify assistance for addressing syntax errors or inconsistencies.

\subsubsection{Debug Agent}
The Debug Agent verifies the functionality and modifies the Verilog code to pass the functionality check from a provided testbench using collaborative Verilog verification tools as shown in Figure~\ref{FlowOverviewFig}(d). 
Given the generated Verilog module from the previous task, the LLM-based Verilog Engineer performs reasoning and interacts with Verilog simulators, as well as the novel {\em AST-WT} through a Thought-Action-Observation process until the generated Verilog code passes the functionality check. 
Figure~\ref{TaskOrientedMultiAIExampleFig} shows an example of the Thought-Action-Observation process of the Verilog engineer fixing functionality issues through reasoning and interaction with {\em Verilog simulator tool} and {\em AST-WT}.

%\vspace{-0.2cm}
\section{Experimental Results} \label{ExperimentalSection}
Our work is implemented in Python and is built on top of the Autogen~\cite{wu2023autogen} multi-AI agent framework. We employ VerilogEval-Human v2~\cite{Batten2024verilogeval2}, which extends the 156 problems of VerilogEval-Human from~\cite{liu2023verilogeval} to specification-to-RTL tasks, as our evaluation benchmark.
\textcolor{black}{We use the same planning, coding, and debugging prompts for these 156 problems.}
%\footnote{\textcolor{black}{We have identified inconsistencies between the reference design module and the specifications in approximately 7\% of the benchmark cases. GPT-4 Turbo without agent-based approach did not demonstrate obvious improvement in these cases with updated specifications, which are detailed in the appendix. We are working with the authors of VerilogEval-Human v2~\cite{Batten2024verilogeval2} to upload the updated specifications to https://github.com/NVlabs/verilog-eval.}}.
To check the functional correctness, the generated Verilog code is tested with the provided golden testbench. 
We measure Verilog functional correctness by running the VerilogCoder once for each problem in the benchmark.

%\textcolor{black}{In the task planner, we extract the circuit signal, transition, and example once, and set max rounds of group chat to 5 for high-level planner agent, TCRG retrieval agent.
% For Verilog code implementation, the max rounds of group chat are set to 20 for code agent, and debug agent.}
Firstly, we demonstrate the Verilog functional correctness of prior works and the proposed VerilogCoder in the Main Results. Next, we conduct an ablation study on the impact of various types of planners and on the effect of using the proposed {\em AST-WT} for specification-to-RTL tasks.

\subsection{Main Results}
\textcolor{black}{We demonstrate the pass-rates of the proposed method and prior works on the VerilogEval-Human v2 benchmark. 
We use OpenAI's GPT-4 Turbo~\cite{GPT4Turbo} and Llama3~\cite{Llama3} as the LLM models for the proposed VerilogCoder (Llama3) and VerilogCoder (GPT-4 Turbo), respectively, in the main experiment.
The temperature and top\_p parameters of the LLMs are set to 0.1 and 1.0, respectively. 
As we are the first to explore using an \textcolor{black}{agentic method} to generate functionally correct Verilog code, we compare the proposed VerilogCoder with recent LLMs using prompt engineering approaches. Table~\ref{MainResultTbl} shows the pass rates for RTL-Coder~\cite{liu2023rtlcoder}, DeepSeek Coder~\cite{guo2024deepseek}, CodeGemma~\cite{Codegemma7b}, CodeLlama~\cite{CodeLlama3}, Llama3~\cite{Llama3}, Mistral Large~\cite{MistralLarge}, GPT-4~\cite{GPT4}, GPT-4 Turbo~\cite{GPT4Turbo}, and the proposed VerilogCoder. 
\textcolor{black}{For a fair comparison, we report the highest pass@1 score across 0-shot, 1-shot, and a sample size ranging from 1 to 20 on the Specification-to-RTL tasks from~\cite{Batten2024verilogeval2}.}
For the VerilogEval-Human v2 benchmark, the proposed VerilogCoder (Llama3) successfully improves the Verilog coding ability of the open-source model and achieves 25.6\% and 7.3\% higher pass rates than Llama3 and GPT-4 Turbo with few-shot and in-context learning techniques~\cite{Batten2024verilogeval2}, respectively.
Moreover, the proposed VerilogCoder (GPT-4 Turbo) not only achieves a 94.2\% pass rate but also outperforms the state-of-the-art recent LLMs GPT-4 and GPT-4 Turbo by 43.6\% and 33.9\%, respectively.}

\textcolor{black}{Here, the average number of group chat rounds for the high-level planner agent and the TCRG retrieval agent is 1.58 and 1.09, respectively. The code agent makes an average of 2.37 Verilog simulator tool calls and 1.37 {\em AST-WT} calls.} \textcolor{black}{The average token count of VerilogCoder is approximately 13$\times$ more than the GPT-4 Turbo baseline method.}

\begin{table}[!t]
\tabcolsep = 1.2pt
\centering
\scriptsize{
\begin{tabular}{|c|c|c|c|c|}
\hline
Method                         & \multicolumn{1}{l|}{Model Size} & \multicolumn{1}{l|}{Model Type} & \multicolumn{1}{l|}{Pass-Rate (\%)} \\ \hline
RTL-Coder                      & 6.7B                            & Open                      & 36.5                                                         \\ 
DeepSeek Coder                 & 6.7B                            & Open                      & 28.2                                                          \\ 
CodeGemma                      & 7B                              & Open                      & 23.1                                                          \\ \hline
DeepSeek Coder                 & 33B                             & Open                      & 37.2                                                           \\ 
CodeLlama                      & 70B                             & Open                      & 41.0                                                          \\ 
Llama 3                        & 70B                             & Open                      & 41.7                                                          \\ \hline
Mistral Large                  & Undisclosed                     & Closed                    & 48.7                                                          \\ 
GPT-4                          & Undisclosed                     & Closed                    & 50.6                                                           \\ 
GPT-4 Trubo                    & Undisclosed                     & Closed                    & 60.3                                                           \\ \hline \hline
{\bf VerilogCoder (Llama3)} & 70B                               & Open                    & {\bf 67.3}                                                             \\ 
{\bf VerilogCoder (GPT-4 Turbo)} & Undisclosed                               & Closed                    & {\bf 94.2}                                                             \\ \hline
\end{tabular}
}
\caption{\textcolor{black}{Pass-rates of recent large language models (i.e., non-agentic method) and the proposed VerilogCoder. 
We run the VerilogCoder once for each problem in the benchmark. The pass-rates of VerilogCoder (agentic method) = \#passed\_case/\#total\_case.
For the pass-rates of recent large language models, we report the best pass@1 score across 0-shot, 1-shot, and sample sizes ranging from 1 to 20 on the specification-to-RTL tasks from~\cite{Batten2024verilogeval2}.}} \label{MainResultTbl}
\vspace{-0.5cm}
\end{table}

\subsection{Ablation Study}
We conducted an ablation study to evaluate the impact of various types of planners, both with and without the proposed {\em AST-based waveform tracing tool}. 
\textcolor{black}{We list two types of planners: (a) {\em Planner1}: A multi-LLM agent consisting of a planner and verilog engineer, and (b) {\em Planner2}: The proposed TCRG based task planner for task-oriented solving.}
%Below, we list the two types of planners included in the study:
%\begin{itemize}
%    \item {\em Planner1}: A multi-LLM agent consisting of a planner and verilog engineer.
%    \item {\em Planner2}: The proposed TCRG based task planner for task-oriented solving.
%\end{itemize}
In {\em Planner1}, given a module description or specification, the planner first decomposes the task into sub-tasks, and the Verilog engineer generates functionally correct Verilog code, including interactions with the provided Verilog verification tools. If syntax or functionality errors occur, the planner debugs and suggests alternative fixes for the Verilog engineer to correct the code. This iterative process between the planner and the Verilog engineer continues until the syntax and functionality are correct or the number of consecutive auto-replies in the group chat exceeds the maximum limit of 100. 

Table~\ref{AblationTbl} shows the pass-rates from the ablation study involving the combinations of {\em Planner1}, {\em Planner2}, and the proposed {\em AST-based waveform tracing tool} on the VerilogEval-Human v2 benchmark. With {\em Planner1}, the {\em AST-WT} achieves a 11.5\% improvement in pass-rate. In contrast, {\em Planner2} without {\em AST-WT} improves by 7.7\% compared to the baseline. Combining {\em Planner2} with {\em AST-WT}, as in the proposed VerilogCoder, significantly improves the pass-rate by 27.5\% compared to the baseline. 

\begin{table}[!t]
% \vspace{-0.2cm}
% \tabcolsep = 2 pt
\centering
\scriptsize{
\begin{tabular}{|c|c|c|}
\hline
               & {\em Planner1}                                                    & {\em Planner2}                                                  \\ \hline
without {\em AST-WT} & \begin{tabular}[c]{@{}c@{}}66.7\%\\ (baseline)\end{tabular} & \begin{tabular}[c]{@{}c@{}}74.4\%\\ (\textcolor{ForestGreen}{7.7\%})\end{tabular}  \\ \hline
with {\em AST-WT}    & \begin{tabular}[c]{@{}c@{}}78.2\%\\ (\textcolor{ForestGreen}{11.5\%})\end{tabular}   & \begin{tabular}[c]{@{}c@{}}94.2\%\\ (\textcolor{ForestGreen}{27.5\%})\end{tabular} \\ \hline
\end{tabular}
}
\caption{Pass-rate (\%) of Ablation study of {\em Planner1} without {\em AST-WT}, {\em Planner1} with {\em AST-WT}, {\em Planner2} without {\em AST-WT}, {\em Planner2} with {\em AST-WT}. AST-WT=AST-based waveform tracing tool. {\em Planner1} without {\em AST-WT} is the baseline, and {\em Planner2} with {\em AST-WT} is the proposed VerilogCoder.} \label{AblationTbl}
\vspace{-0.2cm}
\end{table}

\begin{figure} [!t]
	\centering
	\includegraphics[width=\columnwidth]{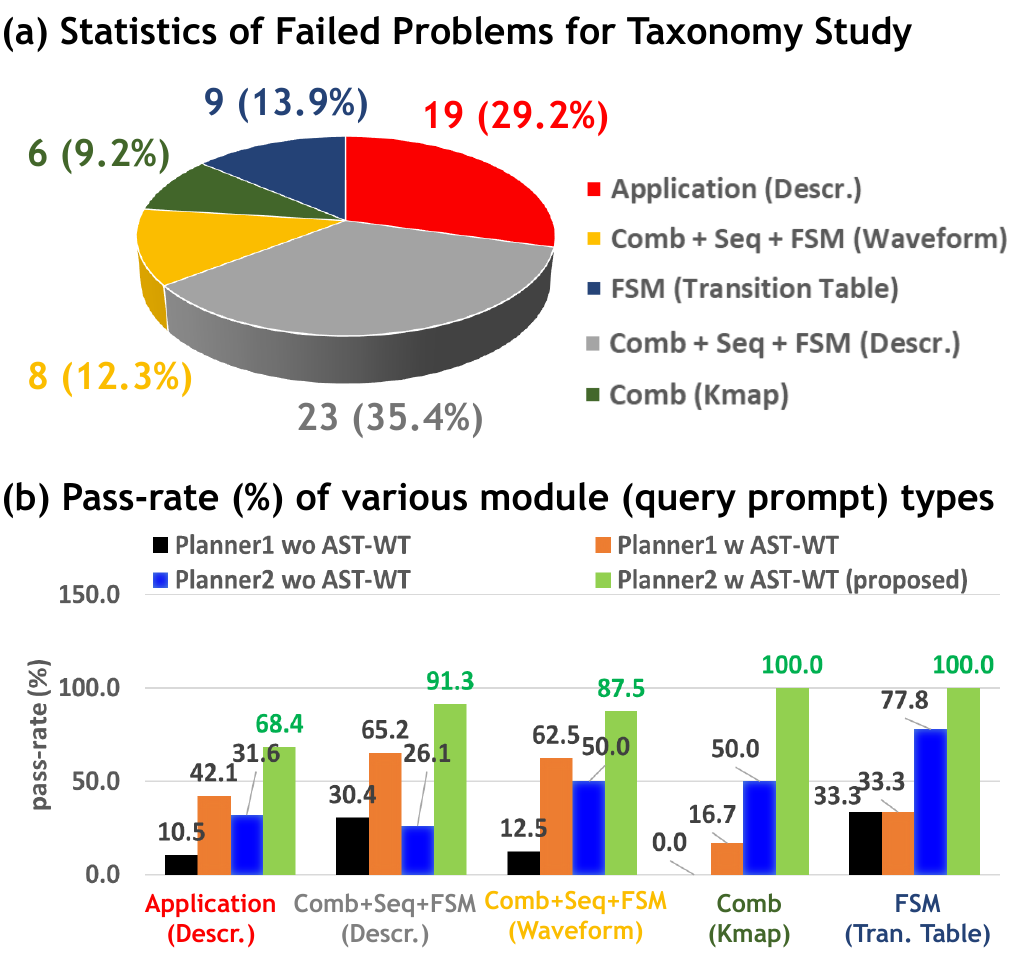}
    % \vspace{-0.7cm}
	\caption{Taxonomy study results. (a) The statistics of extracted failed problems set and the number of problems in each module and query prompt type category. (b) \textcolor{black}{Pass-rate (\%) of each module and query prompt type categories.}}
	\label{TaxonomyDataFig}
    % \vspace{-0.5cm}
\end{figure}

To further investigate the reasons behind the significant improvement in the pass rate of the proposed VerilogCoder, we extract the union set of failed problems from the four combinations and categorize them based on the module and query prompt type of each failed problem for taxonomy study.  
Figure~\ref{TaxonomyDataFig}(a) shows the statistics of the number of each category and their description are listed below.
\begin{itemize}
    \item Application (Descr.): The module is considered for an application (i.e., maze games, lemmings, timer, etc) with descriptions of its functionality in the query prompt.
    \item Comb+Seq+FSM (Descr.): The module is a block of combinational logic, sequential components, or finite state machine (FSM) with descriptions of its connections, and state transitions in the query prompt.
    \item Comb+Seq+FSM (Waveform): The module is a block of combinational logic, sequential components, or FSM with tabular waveform examples in the query prompt.
    \item Comb (Kmap): The module is a block of combinational logic with the Karnaugh map in the query prompt.
    \item FSM (Trans. Table): The module is a FSM block with the state transition table in the query prompt.
\end{itemize}

Figure~\ref{TaxonomyDataFig}(b) shows the pass-rate (\%) of {\em Planner1} without {\em AST-WT}, {\em Planner1} with {\em AST-WT}, {\em Planner2} without {\em AST-WT}, and the proposed method. 
We observe that {\em Planner1} with {\em AST-WT} achieves 10.5\%, 39.1\%, and 12.5\% higher pass-rates on the Application (Descr.), Comb+Seq+FSM (Descr.), and Comb+Seq+FSM (Waveform) categories than {\em Planner2} without {\em AST-WT}, respectively. 
The agent needs {\em AST-WT} to iteratively modify the generated Verilog code, as the indirect transformation from description and waveform to hardware description language may lead to confusion and misleading information. 
On the other hand, {\em Planner2} without {\em AST-WT} outperforms {\em Planner1} with {\em AST-WT} on the Comb (Kmap) and FSM (Trans. Table) tasks by 33.3\% and 44.5\%, respectively. 
This is because the proposed task planner can accurately capture the specified input-output mappings or state transitions in the plan without missing any information, ensuring that the code agent solves the sub-tasks step-by-step. Consequently, with the assistance of the proposed task planner and the {\em AST-based waveform tracing tool}, the proposed VerilogCoder can significantly improve the pass-rate across these types of tasks in the benchmark.

\section{Conclusion and Future Work} \label{ConclusionSection}
Our proposed VerilogCoder \textcolor{black}{has} demonstrated the capability to autonomously write Verilog code and fix syntax and functional errors using the \textcolor{black}{Verilog simulator and the proposed {\em AST-WT}.}
% (i.e., syntax checker, simulator, and waveform tracer). 
%Moreover, we are among the pioneers in utilizing multiple AI agents for autonomous Verilog code completion, including syntax and functional correction. 
\textcolor{black}{The ablation study reveals that the proposed novel TCRG-based task planner and task-oriented solving approach show a 7.7\% improvement in pass-rate.
% particularly for problems described using Kmaps and Transition Tables. 
Additionally, the proposed {\em AST-WT} achieves an 11.5\% improvement in pass-rate.}
%mainly for problems involving Descriptions and Waveforms in the prompt.}
In summary, with the proposed TCRG based task planner and {\em AST-WT}, the proposed Verilogcoder achieves a 33.9\% higher pass-rate compared to the state-of-the-art \textcolor{black}{method.} 

We also believe that important directions for future Verilog agent-based research include: \textcolor{black}{(1) training LLMs with high-quality Verilog code}, (2) improving the generated Verilog code by considering PPA metrics, and (3) incorporating more efficient self-learning techniques and memory systems to enable the agent to accumulate experiences and continuously improve the quality of the generated Verilog code in terms of PPA metrics in the design flow.

\newpage
\bibliography{aaai25}

% \newpage

% \onecolumn
% \includepdf[pages={1-}]{./Figures/Reproducibility_Checklist.pdf}

\end{document}